\documentclass[10pt,twocolumn,letterpaper]{article}

\usepackage{cvpr}
\usepackage{times}
\usepackage{epsfig}
\usepackage{graphicx}
\usepackage{amsmath}
\usepackage{amssymb}

\usepackage{caption}
\usepackage{subcaption}

\usepackage{tabularx}
\usepackage{colortbl}
\usepackage{hhline}

\usepackage[breaklinks=true,bookmarks=false]{hyperref}

\cvprfinalcopy 

\def\cvprPaperID{****} 

\ifcvprfinal\fi

\begin{document}

\title{Teachers' Perception in the Classroom}

\author{\"{O}mer S\"{u}mer$^1$ \qquad Patricia Goldberg$^1$ \qquad Kathleen St\"{u}rmer$^1$\\
Tina Seidel$^3$ \quad Peter Gerjets$^2$ \quad Ulrich Trautwein$^1$ \quad Enkelejda Kasneci$^1$\\
$^1$ University of T\"{u}bingen, Germany\\
$^2$ Leibniz-Institut f\"{u}r Wissensmedien, Germany\\
$^3$ Technical University of Munich, Germany\\
}

\maketitle
\thispagestyle{empty}

\begin{abstract}
    The ability for a teacher to engage all students in active learning processes in classroom constitutes a crucial prerequisite for enhancing students’ achievement. Teachers' attentional processes provide important insights into teachers' ability to focus their attention on relevant information in the complexity of classroom interaction and distribute their attention across students in order to recognize the relevant needs for learning. In this context, mobile eye tracking is an innovative approach within teaching effectiveness research to capture teachers' attentional processes while teaching. However, analyzing mobile eye-tracking data by hand is time consuming and still limited. In this paper, we introduce a new approach to enhance the impact of mobile eye tracking by connecting it with computer vision. In mobile eye tracking videos from an educational study using a standardized small group situation, we apply a state-of-the-art face detector, create face tracklets, and introduce a novel method to cluster faces into the number of identity. Subsequently, teachers' attentional focus is calculated per student during a teaching unit by associating eye tracking fixations and face tracklets. To the best of our knowledge, this is the first work to combine computer vision and mobile eye tracking to model teachers' attention while instructing.
\end{abstract}
\section{Introduction} \label{sec1:introduction}
	
How do teachers manage their classroom? This question is particularly important for efficient classroom management and teacher training. To answer it, various classroom observation techniques are being deployed. Traditionally, approaches to classroom observation, such as teacher instruction and student motivation, have been from student/teacher self-reports and observer reports. However, video and audio recordings from field cameras as well as mobile eye tracking have become increasingly popular in the recent years. Manual annotation of such recorded videos and eye tracking data is very time-consuming and not scalable. In addition, it cannot be easily untangled by crowd-sourcing due to data privacy and the need of expert knowledge.

Machine learning and computer vision, with the advance of deep learning, have progressed remarkably and solved many tasks comparable with or even better than human performance. For example, literature in person detection and identification, pose estimation, classification of social interactions, and facial expressions enables us to understand fine-scale human behaviors by automatically analyzing video and audio data. Human behavior analysis has been applied to various fields, such as pedestrian analysis \cite{Keller:2014}, sports \cite{Bertasius:2017,Felsen:2017}, or affective computing \cite{Valstar:2016}. However, the use of automated methods in educational assessment is not so widespread.  

Previous work in automated classroom behavior analysis concentrate on the activities of students using field cameras or 3D depth sensors and leveraged students' motion statistics, head pose, or gaze \cite{Bidwell:2011,Raca:2015,Venture:2016,Zaletelj:2017:b}. Furthermore, the engagement of students in videos has been studied in educational settings \cite{Whitehill:2014,Bosch:2016,Monkaresi:2017}. 

Students' behaviors are very important to understand the teachers' success in eliciting students' attention and keeping them engaged in learning tasks. However, the view of teachers is an underestimated perspective. How do they divide their attention among students? Do they direct the same amount of attention to all students? When a student raises her or his hands and asks a question, how do they pay attention? Such questions can be answered using mobile eye trackers and egocentric videos which are collected while instructing. Even though there are some previous studies in education sciences, they do not leverage mobile eye tracking data in depth and depend on manual inspection of recorded videos.

In this paper, we propose a framework to combine egocentric videos and gaze information provided by a mobile eye tracker to analyze the teachers' perception in the classroom. Our approach can enhance previous eye tracking-based analysis in education sciences, and also encourages future studies to work with larger sample size by providing in-depth analysis without annotation. We detect all faces in egocentric videos from teachers' eye glasses and create face tracklets from a challenging first person perspective, and eventually associate tracklets to identity. This provides us with two important information: one is whether the teacher is looking at whiteboard/teaching material or student area, and the second is which student is at the center of the teacher's attention at a specific point in time. In this way, we create the temporal statistics of a teacher's perception per student during instruction. As well as per student analysis, we integrate a gender estimation model, as an example of student characteristics, to investigate the relation between the teachers' attentional focus and students' gender  \cite{Einarsson:2002,Dee:2005} in large scale data. Additionally, we propose teachers' movement and view of eye by use of flow information and number of detected faces. 

\section{Related Works}\label{sec2:relatedworks}
	
In this section we address the related works in teacher attention studies using mobile eye tracking (MET), the eye tracking in the domain of Computer Vision, attention analysis in egocentric videos, and face clustering.

\textbf{Mobile eye tracking for teacher's attentional focus.} The first study which links MET and high-inference assessment has been done by Cortina et al. \cite{Cortina:2015}. They used fixation points and manually assigned them to a list of eight standard area of interests (e.g. black board, instructional material, student material, etc.). They investigated the variation of different skills and variables among expert and novice teachers.

Wolff et al. \cite{Wolff:2016} used MET to analyze visual perception of 35 experienced secondary school teachers (experts) and 32 teachers-in-training (novices) in problematic classroom scenarios. Their work is based on Area of Interest (AOI) grid analysis, number of revisits/skips, and verbal data (textometry). The same authors investigated in a follow-up work \cite{Wolff:2017} the differences between expert and novice teacher in the interpretation of problematic classroom events by showing them short recorded videos and asking their thoughts verbally.

McIntyre and Foulsham \cite{McIntyre:2018} did the analysis of teachers' expertise between two cultures, in the UK and Hong Kong among 40 secondary school teachers (20 experts, 20 novices) using scanpath analysis. Scanpath is ``repetitive sequence of saccades and fixations, idiosyncratic to a particular subject [person] and to a particular target pattern''.

In \cite{Sturmer:2017}, on which the paper presented here is based on their recordings, St\"{u}rmer et al. assessed the eye movements of 7 preservice teachers using fixation frequency and fixation duration in standardized instructional situations (M-Teach) \cite{Sturmer:2015} and real classrooms. They studied preschool teachers' focus of attention across pupils and blackboard, however their analysis also requires to predine AOI's by hand in advance.
 
The common point of previous studies in education sciences is that they either depend on predefined AOI's or manually annotated eye tracking output. Furthermore, none of these studies addressed the distribution of teachers' attention among students in an automated fashion. To our knowledge, none of the previous studies on teacher perception and classroom management incorporated MET and CV methodologies in order to interpret attention automatically and in a finer scale.\\
	
\textbf{Eye tracking in Computer Vision.} Looking into the literature, the most common use of eye tracking in CV is in the realm of saliency estimation. Saliency maps mimick our attentional focus when viewing images and are created from the fixation points of at least 20-30 observers in free-viewing or task-based/object search paradigm. Whereas initial bottom-up works in saliency estimation have used local and global image statistics go back to \cite{Treisman:1980,Koch:1987,Itti:1998}, the first model which measures the saliency model against human fixations in free-viewing paradigm was done by Parkhurst and Neibur \cite{Parkhurst:2002}. The most recent state-of-the-art methods are data-driven approaches and borrow learned representations of object recognition tasks on large image datasets and adapt for saliency estimation. 

Besides saliency estimation, eye tracking has been also used in order to improve the performance of various CV tasks such as object classification \cite{Papadopoulos:2014,Sattar:2017:a}, object segmentation \cite{Karthikeyan:2015}, action recognition \cite{Mathe:2015}, zero-shot image classification \cite{Karessli:2017}, or image generation \cite{Sattar:2017:b}.
	
\textbf{Attention in egocentric vision.} The widespread use of mobile devices presents a valuable big data to analyze human attention during specific tasks or daily lives. Egocentric vision is an active field and there have been many works \cite{Jayaraman:2017:a,Jayaraman:2017:b}, however there are only a few studies on gaze and attention analysis. In the realm of finescale attention analysis, particularly using eye tracking, no related work is known.

\begin{figure*}[ht!]
	\begin{center}
		\includegraphics[width=0.9\linewidth]{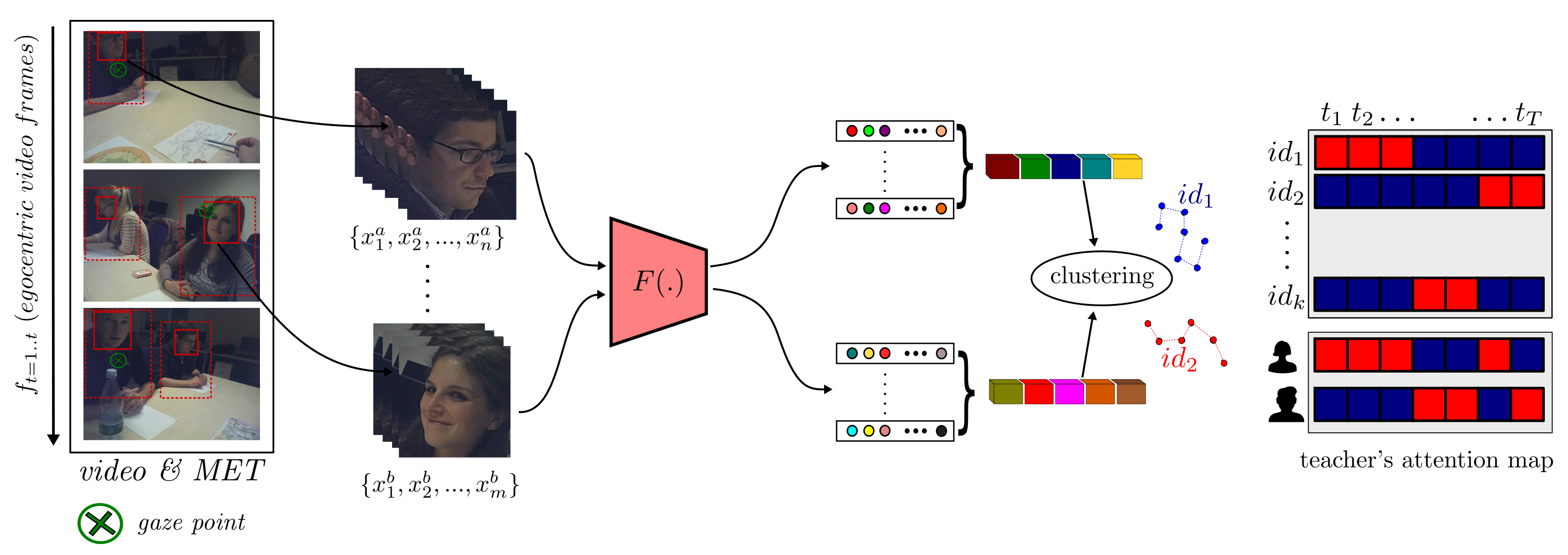}
	\end{center}
	\caption{Teacher's attention mapping workflow. Teachers view and gaze points are recorded by a MET while instructing. In egocentric video sequences, face detection is applied, face tracklets in video are created. Then, features are extracted and aggregated by averaging along the feature dimensions. The aggregated features are clustered. Finally, fixation points are assigned to each identity and attention maps per student identity and gender are created for whole class instruction.}
	\label{fig:01:workflow}
\end{figure*}

Fathi et al. \cite{Fathi:2012:a} analyzed types of social interactions (e.g. dialogue, discussion, monologue) using face detection and tracking in egocentric videos. However, their work does not include eye tracking and gaze estimation for a  finescale analysis of human attention. In another work, the same authors \cite{Fathi:2012:b} used a probabilistic generative model to estimate gaze points and recognize daily activities without eye tracking. Yamada et al. \cite{Yamada:2012} leveraged bottom-up saliency and egomotion information to predict attention (saliency maps) and subsequently assesed the performance of their approach using head-mounted eye trackers. Recently, Steil et al. \cite{Steil:2018} proposed a framework to forecast attentional shift in wearable cameras. However, they exploited several computer vision algorithms as feature representation and used very specialized equipments such as stereo field cameras and head-worn IMU sensors. This make inapplicable in pervasive situations such as educational assessment.

\textbf{Face clustering in videos.} Face clustering is a widely studied topic and applied in still images and video tracklets, which are extracted from movies or TV series \cite{Cinbis:2011,Wu:2013,Shijie:2014}. Many previous studies applied face detection and created low-level tracklets by merging face detections and tracking. In clustering, methods which are based on hand-crafted features exploited additional cues to create must-link and must-not-link constraints to improve representation ability of learned feature space.

The state-of-the-art deep representations are better in dealing with illumination, pose, age changes and partially occlusion and do not require external constraints. Jin et al. \cite{Jin:2017} used deep features and proposed Erdos-Renyi clustering which is based on rank-1 counts along the feature dimension of two compared images and a fixed gallery set. Recently, Nagrani and Zisserman \cite{Nagrani:2018} leveraged videos and voices to identify characters in TV series and movies, but they trained a classifier on cast images from IMDB or fan sites. Particularly the use of voice, which does not happen except for question sessions and training on online cast images, make this approach unsuitable for common educational data. 

Considering previous works in both fields, to the best of our knowledge this is the first work to combine mobile eye tracking and computer vision models to analyze first person social interactions for educational assessment. Furthermore, our approach presents a finescale analysis of teachers' perception in egocentric videos.

\section{Method}\label{sec3:method}
Our goal is to detect all faces which are recorded from teacher's head mounted eye tracking glasses, create face tracklets, and cluster them by identity. Subsequently, we assign eye tracking fixations to student identities and genders when they occur in a small neighborhood of corresponding faces and body regions.  Figure \ref{fig:01:workflow} shows the general workflow of our proposed method. In this section, we will describe our approach to low-level tracklets linking, face representation, features aggregation, clustering, and finally, creation of teachers' attention maps while instructing.

\subsection{Low-level Tracklets Linking} \label{sec3.1:tracklet-linking}

Students mostly sit in the same place during a classroom session, however teachers' attention is shared among whiteboard, teaching material, or a part of the student area. Furthermore, they may also walk around the classroom. Our method first start with face detection and tracklets linking.

Consider there are $T$ video frames. We first apply Single Shot Scale-invariant Face Detector \cite{Zhang:2017} in all frames and detect faces $(x_t^i)_{t=1}^T$, where $i$ is varying number of detected faces. Then, following \cite{Huang:2008}, we created face tracklets $X_K=\{x_1^{i_1},x_2^{i_2},...,x_t^{i_t}\}$ are created using a two-threshold strategy. Between the detections of consecutive frames, affinities are defined as follows:
\begin{equation}
 P_{(i,j)} = A_{loc.}(x_i, x_j) A_{size}(x_i, x_j) A_{app.}(x_i, x_j)
\end{equation}
where $A(.)$ is affinities based on bounding box location, size and appearance. Detected faces between consecutive frames or shots will be associated if their affinity is above a threshold. 

We adopt a low-level association, because clustering based on face tracklets instead of individial detections make subsequent face clustering more robust to outliers. Instead of a two-threshold strategy, which merges safe and reliable short tracklets, a better tracking approach can be considered. However, we observed that egocentric transition between the focuses of attention introduce motion blur and generally faces cannot be detected in succession. A significant proportion of instruction between teachers and students are in the form of dialogue or monologue. Benefiting from this situation, we can mine reliable tracklets, which contain many variations such as pose, facial expression or hand occlusion using position, size, and appearance affinities.

\subsection{Face Representation for Tracklets} \label{sec3.2:face-representation}

Convolutional Neural Networks \cite{Krizhevsky:2014,Simonyan:2014,He:2016,Hu:2017} have become very efficient feature representation for general CV tasks and also performed well in large-scale face verification and identification tasks \cite{Taigman:2014,Liu:2017}. We use and compare these methods as a face descriptor. Patricularly VGG Deep Faces \cite{Parkhi:2015}, SphereFace \cite{Liu:2017} and VGGFace2 \cite{Cao:2017} are among the state-of-the-art methods in face recognition.

Most of these face representations require facial alignment before used in face identification. However, facial keypoint estimation is not very promising in egocentric videos. Furthermore, the image quality, even in the best scenario, is not as good as the datasets where these representations are trained. Additionally by addressing viewpoint and pose variations, we prefer ResNet-50 representation which is trained in VGGFace2 \cite{Cao:2017}.

Using pre-trained networks, we extracted the feature maps of the last fully connected layers before the classifier layer. Then, feature maps are L2 normalized. 

Low-level tracklets $\{X_1,...,X_K\}$ are not of equal length. Thus, we applied element-wise mean aggregation along the feature dimension. Aggregated features are the final descriptor of tracklets and will be further used for clustering.

\subsection{Face Clustering and Attention Maps} \label{sec3.3:clustering-and-attention}

Having video face tracklets, the next step is clustering. In a general image clustering problem, number of clusters and feature representation are first needed to be decided. The number of students is given and we do not need any assumption about number of clusters (identities). When clustering, we do not leverage any must-link or must-not-link constraints, because deep feature representations are robust against various challenges such as pose, viewpoint, occlusion and illumination. 

In teaching videos, we observed that the detections which cannot be associated with others in small temporal neighborhoods either belong to motion blurry frames or occluded. These samples are not representative of their identities and easily be misclassified even by human observers. On the contrary, the temporal tubes which are mined by tracklet linking have dynamics of facial features and more discriminative. For this reason, we applied clustering on only low-level tracklets detected as described in Section \ref{sec3.1:tracklet-linking}.

We used agglomerative hierarchical clustering using Ward's method. First, distance matrix between aggregated features of each tracklets $d_{ij}=d(f(X_i), f(X_j))$. Every point starts in its own cluster and greedily finds and merges closest points until there is only one cluster. Ward's linkage is based on sum-of-squares between clusters, merging cost and in each step, it keeps the merging cost as small as possible.

We train an SVM with radial basis function \cite{Chang:2011} using aggregated tracklet features and their corresponding clustering labels. Subsequently, we predict the category of all non-tracklet detections using this model.

Having clustered tracklets and all detected faces by student identity, we can correspond teacher's focus of attention to students. MET devices deliver egocentric field video and eye tracking data. When acquiring, fixating and tracking visual stimuli, human eyes have voluntary or involuntary movements. Fixations are relatively stable moments between two saccades, fast and simultaneous movements when eye maintained gaze on a location. In attention analysis, only fixation points are used as a significant proximity of visual attention and also work load. 

Eye tracking cameras are generally faster than field cameras. We use a dispersion-based fixation detection method \cite{Santella:2004} and subsequently map fixations to video frames. Then, we assign fixations to the students in case they appear in face region or body of a student. Such attention statistics enable us to better analyze and compare different teachers (i.e. expert and novice) in the same teaching situations. 

\begin{figure}[ht!]
	\centering
		\includegraphics[width=0.95\linewidth]{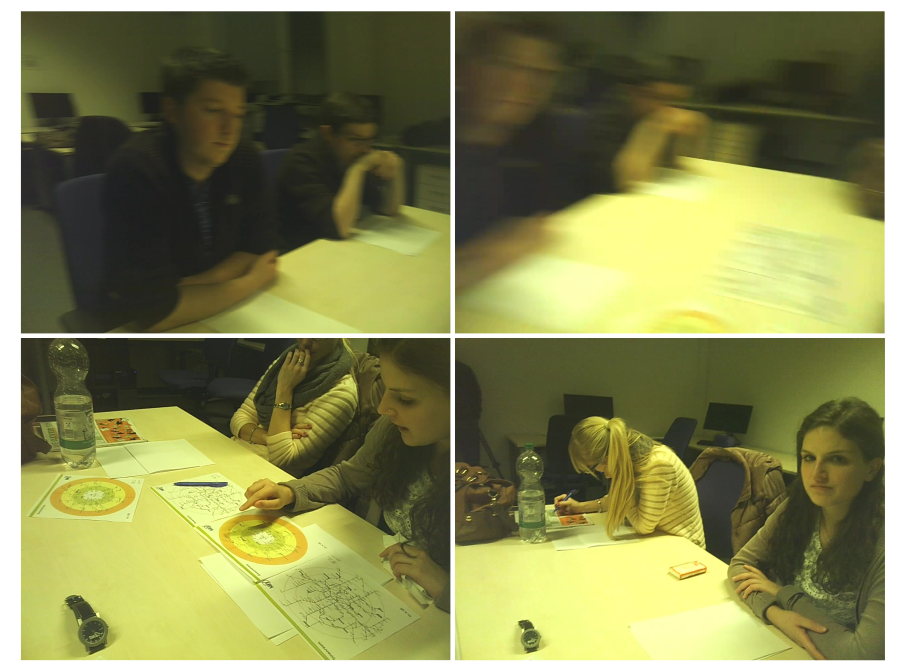}
	\vspace{-0.05cm}
	\caption{ Examples of egocentric vision in M-Teach.}
	\label{fig:data}
\end{figure}
\vspace{-0.25cm}
\section{Experiments}
To validate our approach with real teaching scenarios, we used in a first step the videos excerpts from the study of St\"{u}rmer et al. \cite{Sturmer:2017} in which preservice teachers' taught in standardized teaching settings (M-Teach) with a limited amount of students while wearing mobile eye tracking glasses.

7 M-Teach situations were acquired by mobile eye tracking devices (SMI - SensoMotoric Instruments). Preservice teachers were given a topic (e.g. tactical game, transportation system) with the corresponding teaching material. Based on this material, they made preparation for instructions during 40 minutes, and then taught to a group of four students. In 20-minutes of instruction time, teachers' egocentric videos and gaze points were recorded \cite{Sturmer:2015}.

The recorded videos are in the resolution of 1280$\times$960 and they contain fast camera motion due to first person view. Figure \ref{fig:data} depicts typical example of an egocentric sequence. In this section, our experiments will be done on this representative M-teach video about 15-minute length recorded through the eyes of a preservice teacher. 

\subsection{Feature Representation}\label{Subsec:4.1}

Before analysis of eye tracking data, we need to identify faces of each student detected during the instruction time. To approach this, we used ResNet-50 features. 

\begin{figure}[ht!]
\centering
\includegraphics[width=0.90\linewidth]{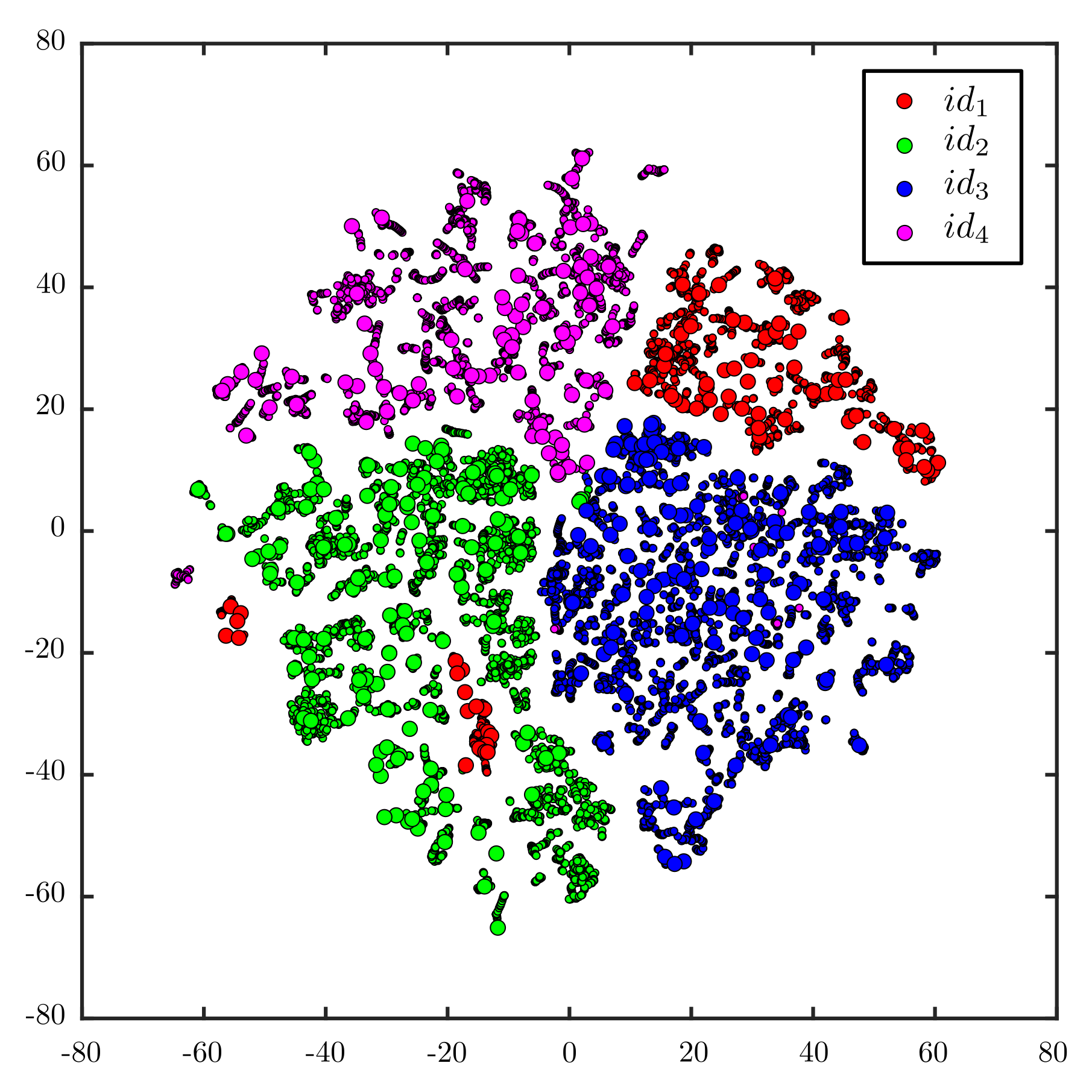}
\caption{t-SNE distribution of face tracklets using ResNet50/VGG2 features.}
\label{fig:tsne}
\end{figure}
\vspace{-0.4cm}
A commonly used face representation, the VGG-Face \cite{Park:2015} network is trained on VGG-Face dataset which contains 2.6 million images. He et al. \cite{He:2016} proposed ``deep residual networks'' and it performed the state-of-the-art on the ImageNet object recognition. Recently, Cao et al. \cite{Cao:2017} collected a new face dataset, VGGFace2 whose images have large variations in pose, lightning, ethnicity, and profession. We prefered ResNet-50 network, which is pretrained on the VGGFace2 dataset. Last feature map before classification layer (2048-dimensional) is l2-normalized and used as feature representation.

\begin{figure}[ht!]
\centering
\includegraphics[width=0.95\linewidth]{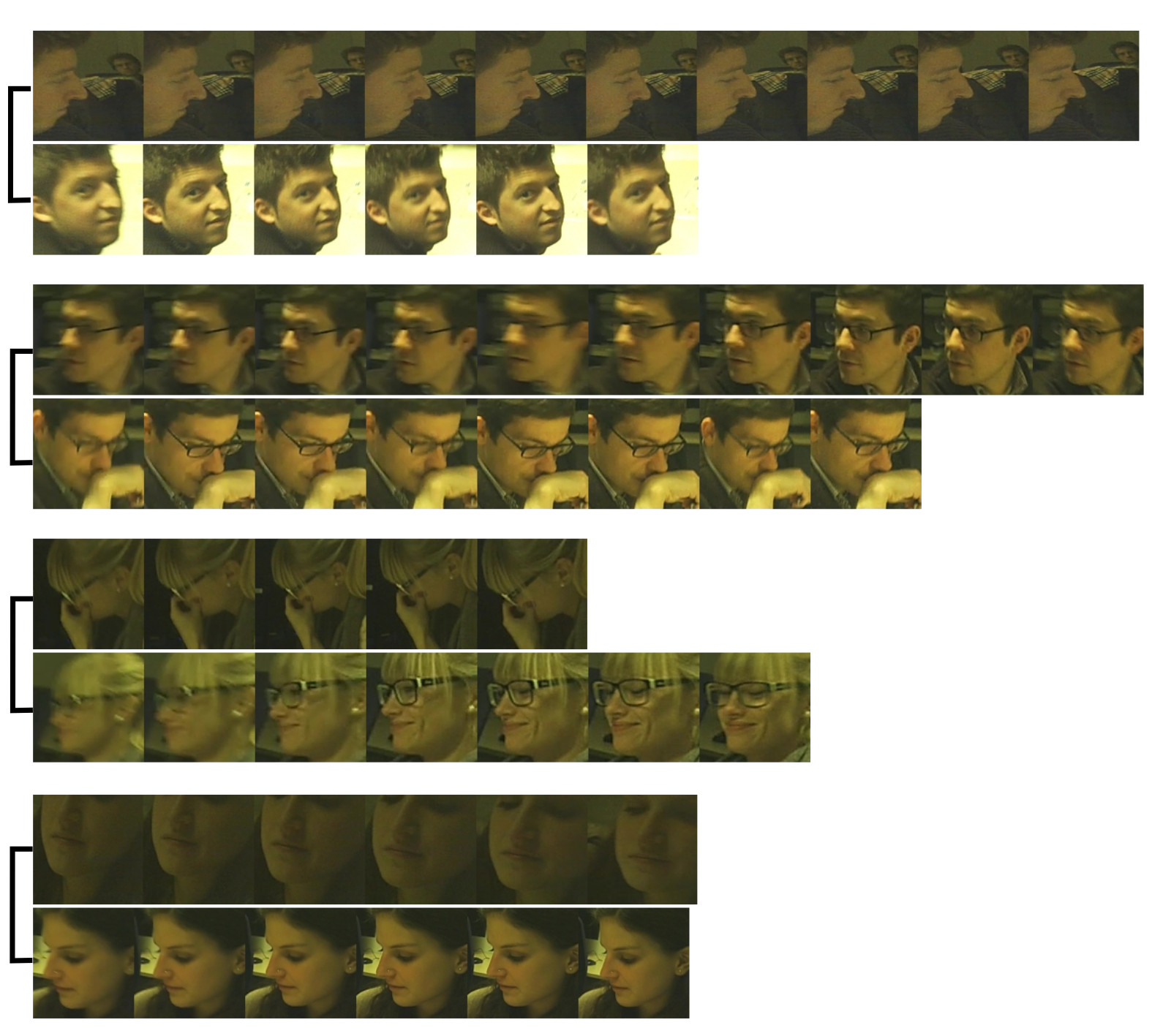}
\caption{Sample face tracklets which are created low-level tracklet linking.}
\label{fig:tracklets}
\end{figure}
\vspace{-0.2cm}
Figure \ref{fig:tsne} shows t-SNE \cite{Maaten:2008} distribution of faces from a M-teach instruction. Big-sized markers represent  face tracklets whose deep features are aggregated by element-wise average, whereas small markers are single faces. Classroom situations are not difficult as general face recognition on unconstrained and web-gathered datasets. However, pose variation is still an issue, because the viewpoint where teachers see the students may greatly vary. Thus, we used ResNet-50 representation which is more discriminative due to the success of residual networks and also more varied training data. Feature aggregation eliminates many outliers and there are only a few misclassified tracklets in one student identity.

\begin{table}[ht!]
\centering
\caption{Confusion matrix of 4-student face clustering}
\label{my-label}
\begin{tabular}{l | l  l  l l}
 & $id_1$  &  $id_2$ &  $id_3$ & $id_4$  \\
 \hline
$id_1$ & \textbf{1897}  & 8 & 13 & 0 \\
$id_2$ & 9 & \textbf{4428} &  28 & 0  \\
$id_3$ & 0 & 13 & \textbf{4558}  & 5  \\
$id_4$ & 0 & 0 & 92 & \textbf{2958}  \\
\end{tabular}
\label{table:confmat}
\end{table}

Figure \ref{fig:tracklets} are the examples of low-level tracklets. It can be seen that some tracklets are blurry, partially detected due to egocentric vision or contain difficult lightning conditions.
\begin{figure*}[ht!]
	\begin{center}
		\includegraphics[width=0.95\linewidth]{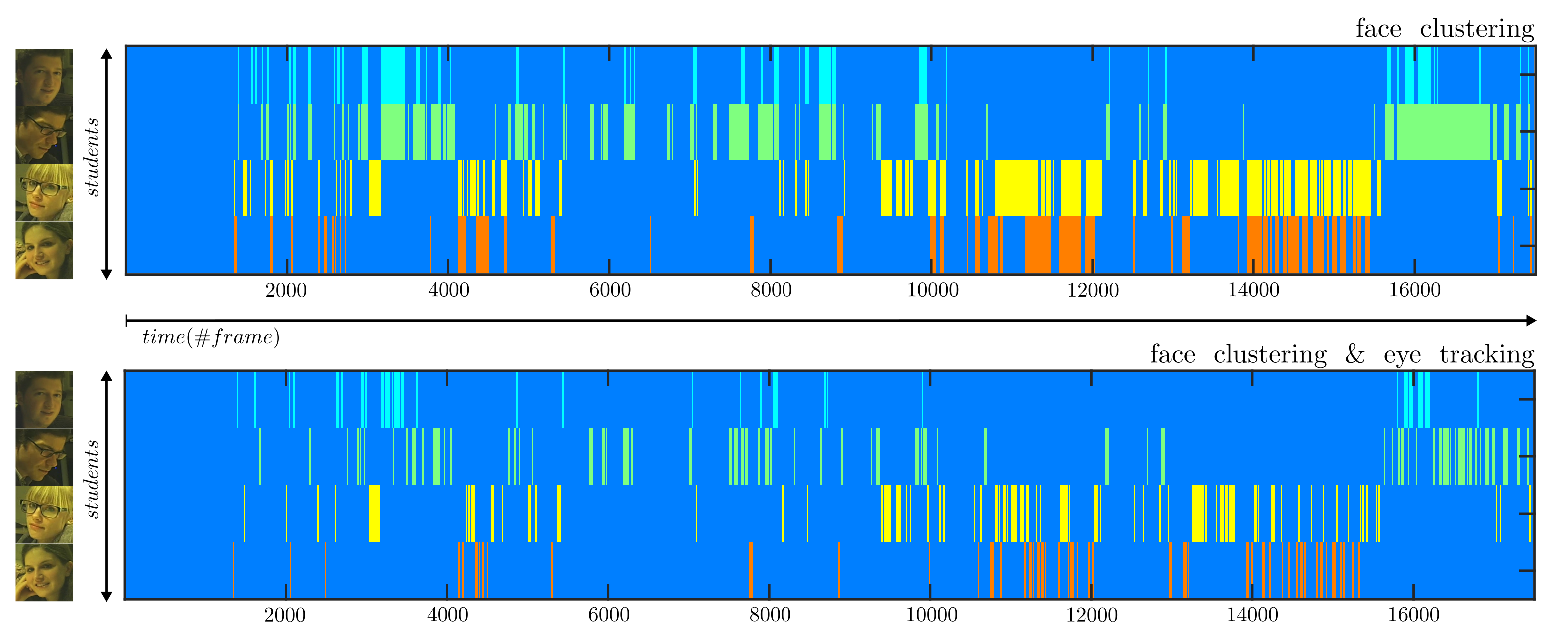}
	\end{center}
	\caption{Attention maps. The results of face clustering during a 15-minute M-teach situation \emph{(above)}, fixation points are assigned to the nearest identity \emph{(below)}.}
	\label{fig:attention}
    \vspace{-0.4cm}
\end{figure*}

We applied agglomerative hierarchical clustering on 2048-dimensional ResNet-50 features. Subsequently, an SVM classifier trained on clustered data in order to assign the detections which cannot be associated with any tracklets. Table \ref{table:confmat} shows the performance of identification  in a 15-minute length M-teach video.

As ResNet-50/VGG2 features are very discriminative even under varied pose, hierarchical clustering without leveraging any constraints performs well. Furthermore, SVM decision on detections which could not linked to any tracklets reduces false classified samples.      

\subsection{Attention Mapping}

After acquiring face tracklets, our final step is to correspond them with eye tracking data. There are four main types of eye movements: saccades, smooth pursuit movements, vergence movements, and vestibulo-ocular movements. Fixations happen between saccades and their lengths vary from 100 to 160 miliseconds. It is generally accepted that the brain processes the visual information during fixation stops. In attention analysis, therefore, mainly fixation events are used. 

We extracted raw gaze points on image coordinates and calculated fixations based on a dispersion-based fixation detection algorithm \cite{Santella:2004}. In our analysis, only fixation events are used.

Figure \ref{fig:attention} depicts a teacher's attentional focus per student during a 15-minute M-teach instruction. First, we show the timeline of frames where each student's face is detected. In this way, we can clearly see which student(s) the teacher interacts in teaching setting. There are moments without any face detection. Teacher either looks at teaching material or explain something on the board by writing. In the second attention map of Figure \ref{fig:attention} represents the distribution of fixation points according to the nearest face.

\begin{figure}[ht!]
	\begin{center}
		\includegraphics[width=\linewidth]{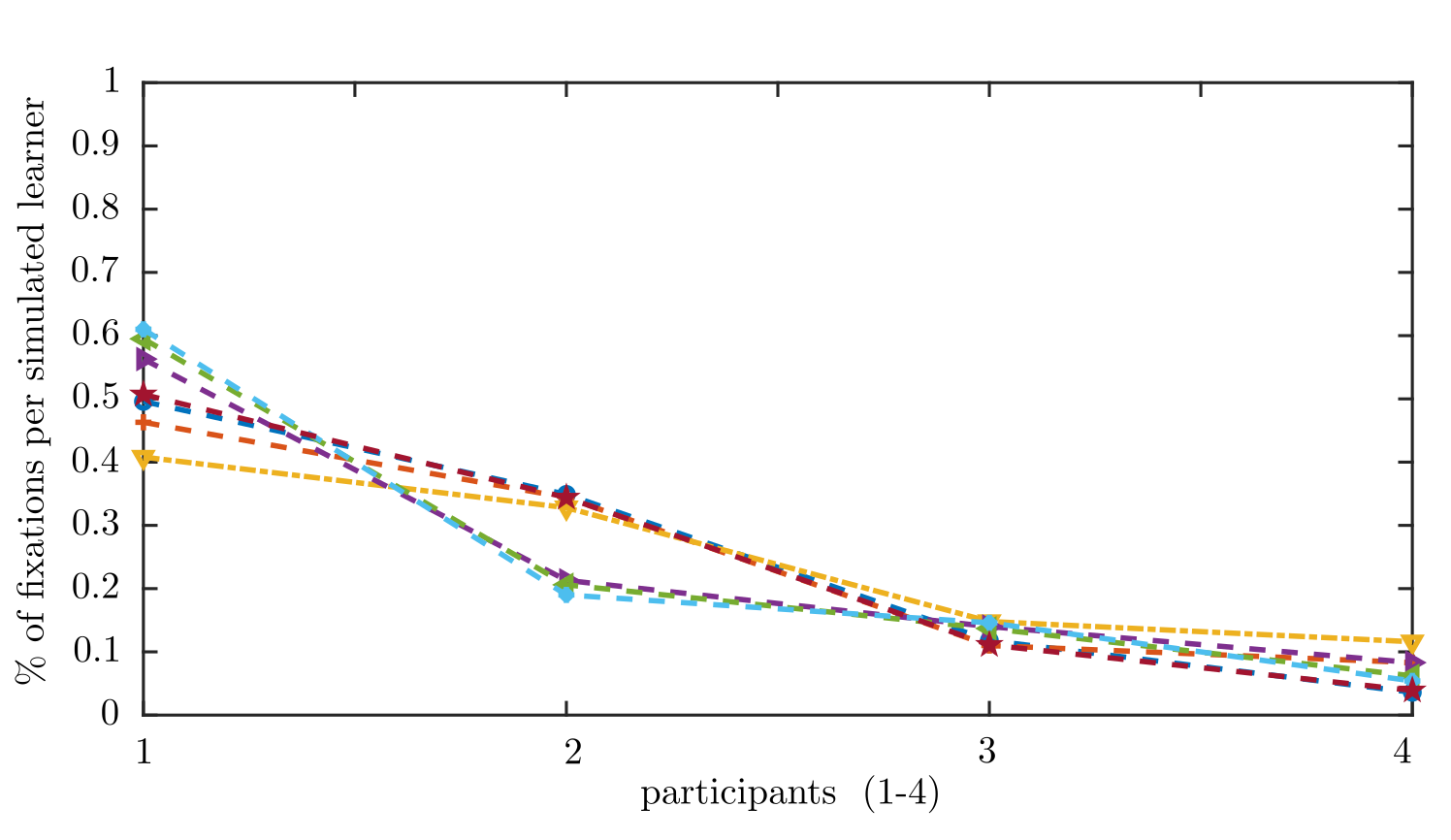}
	\end{center}
    \vspace*{-5mm}
	\caption{Ranked scores for total fixation frequencies per student in 7 M-Teach situations (in descenting order).}
	\label{fig:allvideos}
\end{figure}

After applying our workflow in 7 different M-Teach situations which were captured by different preservice teachers, we created attention maps per teacher. Then, we calculated the percentage of fixations per students from each videos separately. Figure \ref{fig:allvideos} shows that fixation frequencies vary from 40-60\% to 10\%. These results are consistent with S\"{u}rmer et al.'s results \cite{Sturmer:2017} which were based on manually defined AOI's. 

\vspace{-0.1cm}
\subsection{Students' Attributes and Teacher's Attention}
In automated analysis of teacher perception, another interesting question is the relation between teachers' attention and students' attributes, learning characteristics or behavior.
 
\begin{figure*}[t!]
	\begin{center}
		\includegraphics[width=0.95\linewidth]{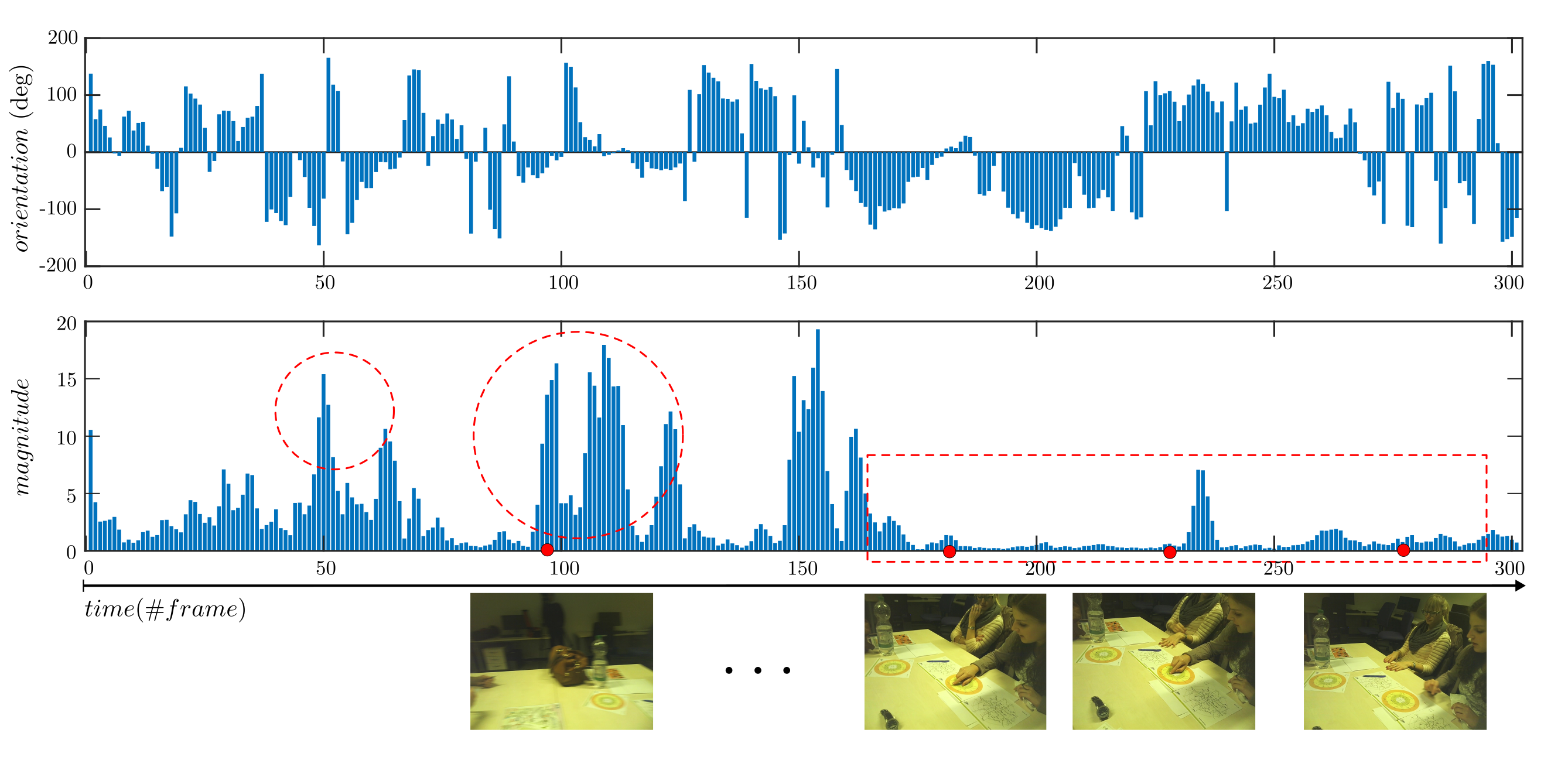}
	\end{center}
	\caption{In a short video snippet, mean magnitude and orientation of otpical flow are shown. Large optical flow displacement indicates that teacher's attentional focus changes. In contrast, long stable areas are indicator of an interaction with a student.}
	\label{fig:flow}
\end{figure*}
 
As an example of these attributes, we exploit gender information. Gender inequality can possibly affect the motivation and performance of students. Thus, our intuition is to extract distribution of teachers' attentional focus according to student gender as well as identity.

Having unique identity tracklets during a video recording of an instruction, one can manually label the gender of each face identity cluster. However, in large scale of data, automatic estimation of gender would be a better approach. Levi and Hassner \cite{Levi:2015} trained an AlexNet \cite{Krizhevsky:2014} on an unconstrained Adience benchmark to estimate age and gender from face images. 

Using face clusters acquired as described in \ref{Subsec:4.1}, we estimated gender of all face images using \cite{Levi:2015} model. For each identity group, we consider the gender estimation of majority as our prediction and subsequently calculate the amount of teacher's eye fixations per student gender while instructing. 
\begin{table}[ht]
\caption{Gender Estimation during an M-teach video} 
\resizebox{\linewidth}{!}{
\begin{tabular}{l l l l } 
\hline 
ID/Gender & \#detections (g.t.) & \#predicted & gender(m/f) \\ [0.5ex] 
\hline 
ID1 (m) & 1918 & 1906 & \textbf{960}/946 \\ 
ID2 (m) & 4465 & 4449 & \textbf{3321}/1128 \\
ID3 (f) & 4576 & 4749 & 879/\textbf{3870} \\
ID4 (f) & 3050 & 2963 & 242/\textbf{2721} \\ [1ex] 
\hline 
\end{tabular}
}
\label{table:gender} 
\end{table}

Table \ref{table:gender} provides the ground truth number of detected faces of four students, the number of predictions from face clustering and gender estimation of all images. It can be seen that gender estimation gives accurate estimation in the majority of predicted clusters. Misclassified proportion is mainly due to blurriness of detected faces. However, we observed that gender estimation performance would be more reliable in longer sequences.

\subsection{Teachers Egocentric Motion as an Additional Descriptor}

As complementary to attentional focus per student identity and gender, another useful cue is teacher's egocentric motion. Some teachers may instruct without any gaze shift by looking at a constant point. Alternatively, they can move very fast among different students, teaching material and board.

Considering that M-teach situation, motion information can also give how frequent teachers' turn between left and right groups of students. For this purpose, we use mean magnitude and orientation of optical flow \cite{Sun:2010}. When using optical flow, we do not intend a high accuracy displacement between all frames of videos. Instead, we aim to spot gaze transitions between students or other source of attention. Figure \ref{fig:flow} shows a typical example of these cases. Mean magnitude of optical flow becomes very large in egocentric transitions, whereas it has comparatively lower values during the dialogue with a student.

Another useful side of optical flow information is to double-check fixation points. Fixation detection methods in eye tracking can spot smooth pursuits or invalid detections as fixation. Optical flow information helps to eliminate falsely classified gaze points. In this way, we can concentrate long and more consistent time intervals in attention analysis. 

\section{Conclusion and Future Directions}

In this study, we showed a workflow which combines face detection, tracking and clustering with eye tracking in egocentric videos during M-teach situations. In previous works in which mobile eye tracking devices were used, association of participant identities and corresponding fixations points have been done by manual processing (i.e. predefined area of interest or labeling).

We have successfully analyzed teacher's attentional focus per student while instructing. Our contribution will facilitate future works which aim at measuring teachers' attentional processes. It can also supplement previously captured mobile eye tracker recordings and provide finer scale attention maps. Furthermore, we showed that attention can be related to students' facial attributes such as gender. Our another contribution is use of flow information to discover teacher's gaze shifts and longer intervals of interaction. It particularly helps to find qualitatively important parts of long recordings.

We also aim to address following improvements on top of our proposed workflow in a future work:

\begin{enumerate}
\item We tested our current approach on eight 15-20 minute length M-teach videos which were recorded from the egocentric perspectives of different preservice teachers. We are planning to integrate our approach to real classroom situation which are taught by expert and novice teachers.
\item Another potential is to leverage students' levels of attention and engagement from facial images and also active speaker detection. In this manner, we can understand why teacher gazes at specific student (i.e. student asks a question or might be engaged/disengaged).
\item Fine-scale face analysis in egocentric cameras is not straightforward. In order to elude the difficulties of egocentric vision, a good solution can be to estimate viewpoint between egocentric and static field camera, and then map eye trackers gaze points into field camera. Thereby, we can exploit better quality images of stable field cameras. 
\end{enumerate}

\paragraph*{Acknowledgements}\mbox{}\\
\"{O}mer S\"{u}mer and Patricia Goldberg are doctoral students at the LEAD Graduate School \& Research Network [GSC1028], funded by the Excellence Initiative of the German federal and state governments. This work is also supported by Leibniz-WissenschaftsCampus T\"{u}bingen ``Cognitive Interfaces''.

\newpage
{\small
\bibliographystyle{ieee}
\bibliography{egbib}
}
	
\end{document}